# Exploring the Role of Electro-Tactile and Kinesthetic Feedback in Telemanipulation Task

Daria Trinitatova, Miguel Altamirano Cabrera, Polina Ponomareva, Aleksey Fedoseev,
and Dzmitry Tsetserukou

*Abstract*—Teleoperation of robotic systems for precise and delicate object grasping requires high-fidelity haptic feedback to obtain comprehensive real-time information about the grasp. In such cases, the most common approach is to use kinesthetic feedback. However, a single contact point information is insufficient to detect the dynamically changing shape of soft objects. This paper proposes a novel telemanipulation system that provides kinesthetic and cutaneous stimuli to the user's hand to achieve accurate liquid dispensing by dexterously manipulating the deformable object (i.e., pipette). The experimental results revealed that the proposed approach to provide the user with multimodal haptic feedback considerably improves the quality of dosing with a remote pipette. Compared with pure visual feedback, the relative dosing error decreased by 66% and task execution time decreased by 18% when users manipulated the deformable pipette with a multimodal haptic interface in combination with visual feedback. The proposed technology can be potentially implemented in delicate dosing procedures during the antibody tests for COVID-19, chemical experiments, operation with organic materials, and telesurgery.

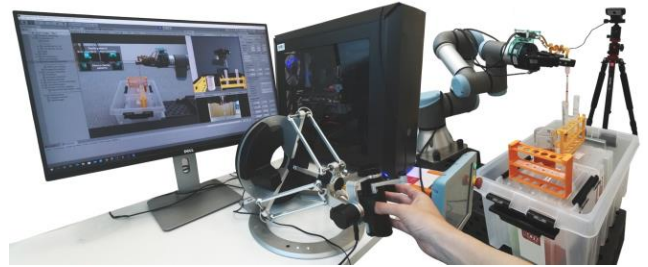

(a) The operator manipulates the plastic pipette with a robotic arm.

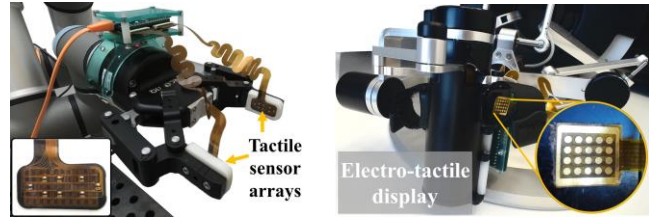

(b) Robotic gripper equipped with high-density tactile sensors.

(c) Electro-tactile display embedded to the handle of the Omega.7 haptic device.

Fig. 1. The developed telemanipulation system. The object shape is captured by the high-fidelity tactile sensor arrays during the grasping and delivered to the operator through the electro-tactile display.

## I. INTRODUCTION

Nowadays, more and more systems and operations with humans in the loop are being switched to autonomous or remotely controlled modes for various reasons, such as the mobility of systems, the convenience of monitoring from anywhere globally, and restrictions related to the epidemiological situation. Most modern robotic systems operate autonomously without requiring human intervention [1], [2]. For example, a fully autonomous robot for chemical manipulation was developed by Burger et al. [3]. This mobile platform, driven by a batched Bayesian search algorithm, can perform a high number of experiments per day to identify photocatalyst mixtures. However, the presence of a human operator is still required to develop a more adjustable system for a nondetermined scenario, such as conducting medical research and operations.

Telemanipulation of robotic systems allows the operator to perform the most complex operations from anywhere in the world with the proper equipment in the operating room. The effectiveness of human interaction with a remote environment without direct physical contact is one of the critical features for the development of such systems. The user immersion in the teleoperation or telexistence systems is fundamental due to the sensation perceived from the environment and the response that the users have to send through the system. The use of haptic feedback during teleoperation provides many benefits to improve task performance, such as regulation of output forces and the operator's situational awareness (e.g., collision avoidance) [4]. Many research works were devoted to the application of haptic feedback for teleoperated robotic surgery which requires high-precision operations [5]. However, a number of routine operations and medical tests are still performed by human personnel manually. In this work, we consider the manipulation of laboratory instruments such as pipettes and test tubes with a robotic arm for remote testing for COVID-19 and other infections as the main scenario.

This paper proposes a robotic telemanipulation system with a multimodal haptic feedback system that combines kinesthetic and electro-tactile stimuli for precise remote dosing of liquids (Fig. 1). The Omega.7 haptic display with 7-DoF and embedded high-fidelity electro-tactile displays provide feedback on the applied force and shape of the contact surfaces of the grasped object. The application of multimodal haptic feedback to the user is expected to improve the quality of telemanipulation allowing to render the contact surfaces and grasp the objects without damaging

The authors are with the Intelligent Space Robotics Laboratory, Center for Digital Engineering, Skolkovo Institute of Science and Technology (Skoltech), 121205 Moscow, Russia. {daria.trinitatova, miguel.altamirano, polina.ponomareva, aleksey.fedoseev, d.tsetserukou}@skoltech.ru

them. The experimental part was designed to investigate the influence of multimodal haptic feedback on the accuracy of manipulations with a deformable object, namely a plastic pipette, in the task of dispensing a liquid.

## II. RELATED WORKS

Nowadays, teleoperation and telepresence are serving as one of the most natural applications of haptics. However, the use of effective combinations of haptic stimuli remain the subject of active research. In the work of Chen et al. [6], the kinesthetic feedback from Geomagic Touch haptic device is used to provide the operator with information about the actual tracking error of the robotic arm during the teleoperation process. Xu et al. [7] introduced a non-delayed kinesthetic rendering with guaranteed system stability through local model-mediated teleoperation architecture. However, in these studies, the effectiveness of force feedback when manipulating small objects is rarely considered and requires further evaluation.

Electro-tactile stimulation has great potential for applications in teleoperation tasks because it can render high-resolution information about grasped surfaces (e.g., texture, shape). Yem et al. [8] studied the effect of electrical stimulation on perception of softness/hardness and stickiness of a virtual object using the electrode arrays placed on the fingertip and on the back of the hand. In the work Peruzzini et al. [9] the utilizing of electrostimulation to simulate the roughness, slickness, and texture coarseness of materials, such as wood, paper, or textile fabric on the fingertip was presented. Hummel et al. [10] introduced a haptic device for multi-finger stimulation with electrotactile feedback. Each electrotactile tactor comprised an array of 8 electrodes. The evaluation user study revealed that the use of electrotactile feedback improved performance in object manipulation and grasping tasks while reducing the workload demands. Pamungkas et al. [11] proposed a teleoperation system of robotic arm with electrotactile feedback generated on the back of the hand using TENS electrodes.

A better dexterous telemanipulation can be achieved by providing the operator with multimodal high-fidelity haptic feedback. Sagardia et al. [12] presented a virtual reality framework with multimodal haptic feedback for simulation and training telerobotic on-orbit servicing tasks. The haptic feedback system can provide force feedback using bimanual grounded haptic device HUG, and vibrotactile and electrotactile feedback at the elbow and the fingertips of the user using portable devices. Pacchierotti et al. [13] developed a teleoperation system that generates kinesthetic and vibrotactile feedback through a pen-shaped handle. The experimental evaluation of this device suggested that the proposed system would help to decrease the robot's tool targeting and orientational errors. Earlier, Sarakoglou et al. [14] presented model-based tactile feedback where a compact 4x4 pin matrix was mounted on a force feedback master device. Kuling et al. [15] studied the role of haptic feedback (direct force feedback and vibration feedback) in the Box &

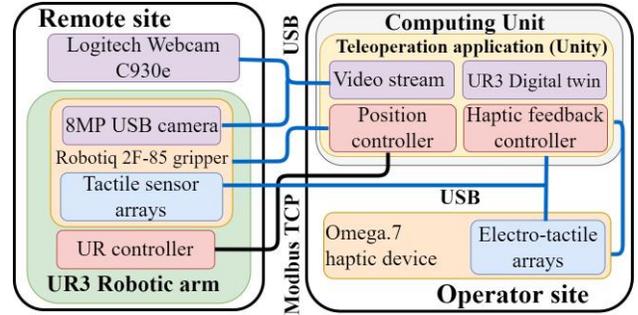

Fig. 2. The architecture of the developed teleoperation system.

Blocks Task. The results of the study did not reveal a user preference for a particular tactile feedback. Aggravi et al. [16] proposed a teleoperation system for flexible needle insertion in soft tissues, which provides a user with kinesthetic and vibrotactile feedback. The experimental evaluation showed that the user performance was better with a combination of kinesthetic and vibrotactile feedback compared to employing only one type of feedback.

However, there is still a demand for a robotic system that can detect the contact force distribution at the tips of the gripper and transmit high-fidelity sensations to the operator's hand to achieve dexterous manipulation of objects, especially in the case of soft or deformable objects.

## III. SYSTEM OVERVIEW

### A. System Architecture

In this work, we propose a telemanipulation system with high-fidelity tactile sensors that provides multimodal haptic feedback to the user's hand. The proposed system consists of a robotic manipulator (UR3) with 6 degrees of freedom equipped with a 2-finger gripper (Robotiq 2F-85), a digital 8-megapixel on-gripper camera with a 75° field of view (FoV), and high-fidelity tactile sensors mounted on the gripper. The operator controls the robot using the Omega.7 desktop haptic device with two embedded electro-tactile displays on the handle. As a visual feedback channel for the operator, we use a hybrid environment composing real-time video streams accompanied by a digital twin of the robot. Visual feedback from the remote environment includes a real-time video stream from the on-gripper camera and the isometric view camera. The overall scheme of the developed system is presented in Fig. 2.

### B. Control system of the robot

The motion of the robot's Tool Center Point (TCP) is controlled by moving the handle of the haptic device from its center position. Three translational axes of the robot TCP correspond to three translational axes of the haptic device, and three rotational axes of the robot TCP are controlled by the handle tilt. The workspace configuration of the haptic device (ø 160 $mm$ x $L$ 110 $mm$) and the robot (R = 500 $mm$) differs from each other. Therefore, the developed control framework performs the workspace scaling from the haptic device to the robot with a scaling factor of x1–x5. The movement of the robot TCP along one or two translational axes, as well as TCP rotation could be locked to increase the accuracy and speed of the robot manipulation. The control signal for TCP movement and the gripper control command are generated from the haptic device and transmitted to the control framework via TCP/IP protocol. The PID controller was developed and applied to provide a stable and precise robot motion. Feedback from the real robot and gripper is applied in each loop to adjust the digital twin state in real-time.

*C. Visual Channel*

The use of visual feedback from the remote environment is the most natural for the operator since it does not require any training and additional equipment aside from the regular camera installation. However, the most significant disadvantage of this approach is the inaccuracy of information caused by the optical distortion and camera occlusion, which prevents the operator from accessing the overall state of the system. To eliminate this disadvantage, it is proposed to use two visual feedback channels.

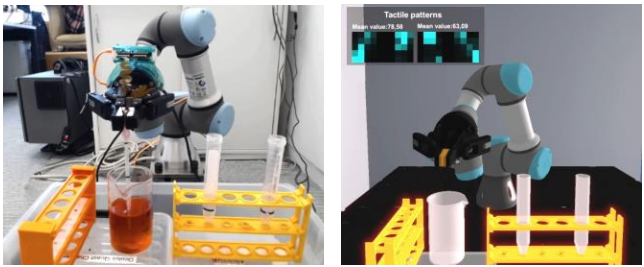

(a) Robotic arm during teleoperation. (b) Unity3D interface displaying a digital twin for visual feedback.

Fig. 3. Robotic manipulator UR3 with its digital twin.

The first visual channel is presented by the digital twin interface of the UR3 robot based on the Unity3D Engine, which was introduced previously in [17]. The digital twin position is based on the real robot feedback, thus providing the information of the real robot within 0.1 $mm$ precision and 50 $Hz$ frequency. The virtual environment does not fully display the remote environment, being able to render only the predefined static obstacles. The key feature of the virtual interface is that it can present the current state of the robot from different perspectives without any complex camera setup (Fig. 3 (b)). Additionally, the digital twin supports the users with visual information about the gripper opening and contact patterns obtained from tactile sensor arrays for convenient control of the current state of the grasped object.

To compensate the lack of dynamic obstacle information, the second visual channel provides feedback from two cameras. The first camera is fastened on the Robotiq gripper and the second camera (Logitech Webcam C930e with 90° diagonal FoV) is located next to the robot, allowing the operator to access the remote environment in real-time.

*D. Embedded tactile sensors*

The 2-finger robotic gripper is equipped with high-density tactile sensor arrays (Fig. 1 (b)). These tactile sensors scan the contact surface, measuring the internal force and pressure applied to the grasped objects during the manipulation process. Both tactile sensor arrays consist of 5 × 10 independent electrodes with high sensitivity, described in [18]. Thus, the robot detects the pressure applied to the object grasped by the gripper with a resolution of 100 points (50 points per finger). The sensing frequency is 120 $Hz$. The range of force detection per point is $1-9$ $N$.

*E. Multimodal haptic feedback*

The proposed telemanipulation system allows receiving high-fidelity tactile feedback from the environment using the high-density tactile sensors embedded in the robotic gripper, allowing for precise operations with target objects. The obtained information is transmitted to the operator in the form of haptic feedback. The user is provided with both kinesthetic and tactile feedback. As a result, the operator can feel both the grasping force applied to the object by a robotic gripper and the detailed information about the force distribution, i.e., the applied pressure at the contact surfaces represented by electro-tactile patterns.

*1) Kinesthetic stimulation:* The kinesthetic channel is composed of the tactile sensing of the robot and the haptic display Omega.7 generated the force feedback. Kinesthetic feedback is provided to the user with the gripper of Omega.7 haptic device. The force feedback is controlled by the PID regulator with the proportional component and calculated as the average force from pressure sensors summarized with the difference between control input from the gripper of haptic device and robot gripper feedback sent to the framework:

$$F_{omega} = \frac{\Sigma F_{tactile}}{100} \cdot (P_{contact} - P_{current}), \quad (1)$$

where $F_{omega}$ is the grasping force generated by the gripper of Omega.7 haptic device, $F_{tactile}$ is data from each tactile sensor arrays, $P_{contact}$ is the position of first contact with the object, $P_{current}$ is current positional feedback from the gripper.

*2) Electro-tactile stimulation:* The electro-tactile feedback is generated using two electro-tactile displays [18] embedded on the handle of the desktop haptic device to deliver the tactile patterns to the thumb and index distal phalanx (finger tips) of the operator (Fig. 1 (c)). The control framework processes the data from each tactile sensor array to provide the perception of object shape using electrode arrays. The size of display matrix is 4 × 5 electrodes for each array. Thus, we resize the sensor data array (5×10 cells), reducing it to the electrode array size (5×4 cells). To downsize the tactile information, we applied a bicubic interpolation algorithm to obtain smooth output array which represents a tactile pattern that retains most of the details of the original contact surfaces. As a result, the operator can feel the surface of the object grasped by the robot.

## IV. USER STUDY

To evaluate the proposed telemanipulation system we elaborated a two-stage experimental scenario for dispensing liquid with a plastic pipette. At the first stage, we investigated the role of each haptic feedback channel for the proposed system during the liquid dispensing operation with semi-autonomous teleoperation system, where the positions of the robot were predefined. At the second stage, we proposed the participants to perform the same operation with a fully teleoperated system.

### A. Exploring the role of haptic feedback in liquid dosing task

The purpose of the experiment was to compare user performance when dispensing liquid using the robotic arm with a pipette in the presence of three different types of feedback and their combination. The haptic feedback provided to the user was presented with three methods of contact force rendering, namely grasping force simulation with Omega.7 haptic device, distributed force simulation with the electro-tactile display, and a combination of the both methods. For this experiment, the UR3 robot has predefined positions, and the robotic gripper was controlled by the operator with an Omega.7 haptic device. Two test tubes with volume markers and water container were placed in front of the manipulator (Fig. 4).

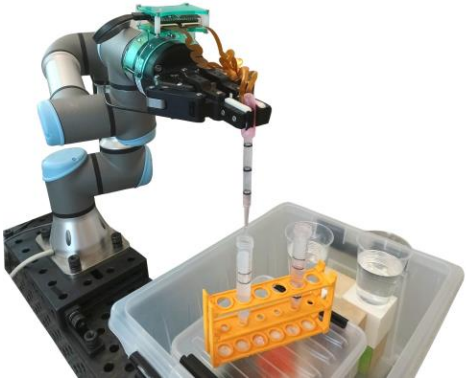

Fig. 4. Experimental setup with a remote robot UR3 for dispensing liquid into two markered test tubes.

*1) Participants:* Eight subjects (7 males and 1 female) volunteered to participate in the experiment. The average participant age was 24.3 (SD=1.66), with a range of 23–31. The participants were informed about the experiments and signed to the consent form.

*2) Experimental procedure:* At the beginning of the experiment, the robot TCP with the pipette grasped by the gripper was positioned above the container. Each participant had to fill a pipette with water and, when the robot TCP reaches a position above the first test tube, add 2 *ml* of liquid to it. Then they were asked to repeat this procedure with the second test tube. In this experiment, each participant carried out the task in the presence of four types of feedback: Visual feedback only (V), Visual and Force feedback (V+F), Visual and Electro-tactile feedback (V+E), and the combination of all the feedback channels described above (V+F+E). The type of feedback changed for each trial in random order for each participant. Before the experiment, a training session was conducted where the each type of feedback was explained and demonstrated to the users.

Two metrics were chosen as the main parameters affecting the teleoperation efficiency, namely accuracy of liquid dosing and time of dosing operation.

*3) Experimental Results:* The experimental results revealed that the presence of haptic feedback reduces the dosing error (Fig. 5 (a)). Most significantly, the mean dosing error in the case of multimodal haptic feedback decreased by 66% relatively to visual feedback mode.

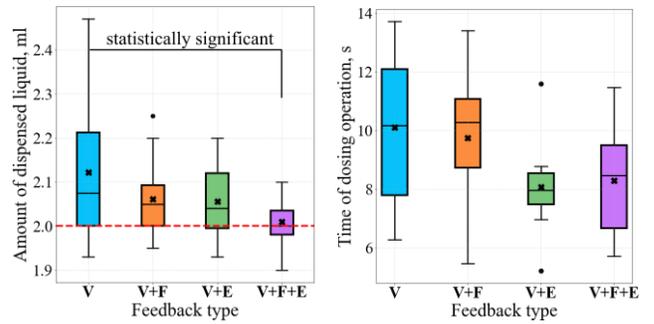

(a) The distribution of liquid dosing for different types of feedback.
(b) Distribution of liquid dosing time for different types of feedback.

Fig. 5. Comparison of amount of dispensed liquid and time of liquid dosing operation for different types of feedback. (a) The red dashed line represents the required volume of 2 *ml* for liquid dispensing. (a)-(b) Crosses mark mean values.

During the control mode with the presence of kinesthetic feedback (V+F), the error decreased by 38% relatively to pure visual mode. During the task performed with electro-tactile feedback only (V+E), the error decreased by 47% on average compared to the pure visual mode, proving the significance of distributed force component in multimodal haptic feedback. However, in this case, the dispersion of error is higher than in the case of multimodal haptic feedback, and a tendency to overdose was observed.

To evaluate the statistical significance of differences in the user accuracy of liquid dosing for 4 feedback types, we analyzed the results using single factor repeated-measures ANOVA, with a chosen significance level of $\alpha < 0.05$. To examine the assumption of a normal distribution of the data, the Shapiro-Wilk was performed and did not show evidence of non-normality (V: $W = 0.9$, $p-val = 0.09$; V + F: $W = 0.92$, $p-val = 0.15$; V + E: $W = 0.94$, $p-val = 0.39$; and V + F + E: $W = 0.92$, $p-val = 0.18$. According to the ANOVA results, there is a statistically significant difference in the performance of the users, $F_{(3,21)} = 3.42$, $p = 0.03$, $\eta_p^2 = 0.33, 90\%$, $\omega_p^2 = 0.10$ suggesting a large effect on the application of the different feedback types on the accuracy of liquid dosing. The paired t-tests with one- step Bonferroni correction showed statistically significant differences between the system with only Visual feedback (V) and the multimodal feedback system (V+F+E) ($p = 0.07 < 0.1$, $Hedges g = 0.93$). The open-source statistical package Pingouin was used for the statistical analysis.

In terms of liquid dosing time, the tasks with electro-tactile and multimodal haptic feedback were performed the most efficiently (Fig. 5 (b)). The task was completed 20% faster for V+E feedback and 18 % faster for V+F+E feedback compared to pure visual feedback. However, the difference in liquid dosing time with different types of feedback is statistically insignificant ($F_{(3,21)} = 2.22$, $p = 0.12 > 0.05$) according to one-way repeated-measures ANOVA.

*B. Performing a liquid dispensing operation in a fully tele-operated mode*

The experimental task was to perform a liquid dosing operation using the developed system with multimodal haptic feedback in the full teleoperation mode. The UR3 robot and the robotic gripper was controlled by the user with an Omega.7 haptic device with x2 scaling mode. The experimental setup included a tube rack with a pipette for manipulation, a glass beaker with colored water and second tube rack with test tubes with visual markers for required dosing volume (Fig. 3 (a)).

*1) Participants:* Twelve participants (9 males and 3 females) volunteered to take part in the experiment. The average participant age was 24.2 (SD=1.82), with a range of 22–28. Eight participants reported having some experience in teleoperation of robotic manipulator, two participants have never operated robots, and two participants have a regulary work with robotic manipulators. None of the participants had prior experience with the proposed telemanipulation system.

*2) Experimental Procedure:* The experimental task was to grab a pipette (1.5 *ml* transfer pipette) from a tube rack, fill it with water from the glass beaker, and then pour the liquid into one of the test tubes. The required amount of liquid to dispense was 2 *ml*, as in the first experiment. Before the experiment, each participant had a short training session to demonstrate the procedure and the control principle of the robot. For each participant, we measured task execution time and the accuracy of liquid dosing. In addition, after completing the task, we asked participants to evaluate the system with NASA TLX questionnaire using a seven-point Likert scale [19].

*3) Experimental Results:* The results of the experiment showed that the average task execution time was about eight and a half minutes (Table I). Participants performed the liquid dosing procedure with the pipette an average of two times to reach the required volume in the test tube, which required delicate manipulations. In terms of liquid dosing accuracy, participants showed comparable results to the first experiment with semi-autonomous control. The average amount of dispensed liquid was 2.07 *ml*, SD=0.25 *ml*. The mean absolute error comprised about 9% of the required volume of 2 *ml*.

Fig. 6 shows the results of the user assessment of the proposed telemanipulation system. The participants highly assessed their performance ($\mu = 2.4$, SD=0.9) with a low level of frustration during the task ($\mu = 2.4$, SD=1.15). The task execution through developed teleoperation system did not require special mental ($\mu = 3.9$, SD=1.51) and physical ($\mu = 3.9$, SD=1.73) costs as well as putting a lot of effort ($\mu = 3.6$, SD=1.55) from the users.

TABLE I
STATISTICS OF TASK EXECUTION TIME AND AMOUNT OF LIQUID DISPENSED

|  | Min | Max | Mean | SD |
|---|---|---|---|---|
| Dispensed liquid, *ml* | 1.8 | 2.5 | 2.07 | 0.25 |
| Task execution time, *s* | 313.5 | 688.2 | 512.6 | 313.5 |

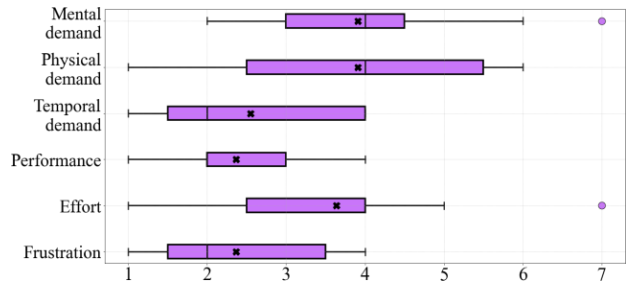

Fig. 6. Evaluation of the participant's experience in the form of a 7-point Likert scale. Crosses mark mean values.

Fig. 7 represents the example of obtained contact patterns from tactile sensors when grasping the pipette at different positions for different grasping states: at the moment of the first touch of sensors with the pipette, in the intervening state of pipette squeezing, and at the moment of maximum gripper closure. The mapped data transmitted to the electrode array helps the operator to precisely distinguish between the contact surfaces at different grasping states, which combined with force feedback, allowed accurate liquid dosing with less pipette compression.

## V. CONCLUSIONS AND FUTURE WORK

In this study, we evaluated the effect of multimodal haptic feedback for the developed telemanipulation system. The experimental results showed that the proposed approach to use a combination of haptic channels (kinesthetic and electro-tactile) allowed operators to increase accuracy (by 66%) and decrease time (by 18%) of liquid dosing task compared to pure visual feedback mode. Overall, users demonstrated better performance using multimodal haptic feedback system than when using each haptic channel separately in combination with visual feedback. In addition, using the proposed telemanipulation system the users showed good accuracy (with an average absolute error of 9% of 2 *ml*) in liquid dosing task during a fully teleoperated mode. Therefore, the proposed system can be potentially implemented in tasks requiring high-fidelity manipulation with small soft objects, such as antibody tests in laboratories, manipulation with chemicals, or cooking.

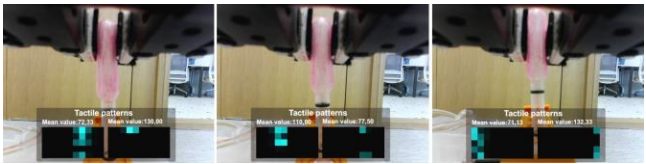
(a) The moment of first contact of the gripper with the pipette.

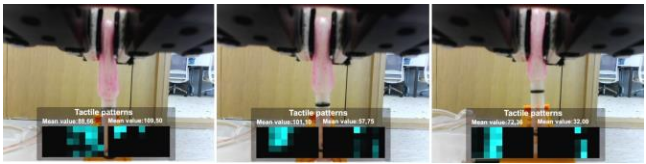
(b) An intermediate state of squeezing the pipette.

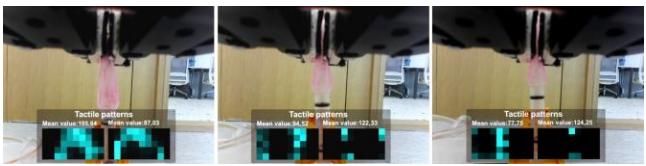
(c) The moment of maximum compression of the pipette by the gripper.

Fig. 7. The pipette grasping at different gripper positions. At the bottom of each image the visualisation of contact patterns from the left and right tactile sensor arrays are presented.

In future work, we are going to use deep learning techniques for the development of the perception system allowing to recognize the properties of the grasped object, as was introduced in [20]. In particular, the detection of the pipette orientation during grasping will allow the user to adjust the position of the gripper to align the pipette vertically [21]. In addition, we will conduct experiments with a larger number of objects that differ in softness, shape and texture.

ACKNOWLEDGEMENTS

The reported study was funded by RFBR according to the research project No. 20-38-90294.

The authors would like to thank Jonathan Tirado for helping with setting up the the tactile sensors and electro-tactile displays.